\documentclass[3p,review,10pt]{elsarticle}

\makeatletter
\def\ps@pprintTitle{%
 \let\@oddhead\@empty
 \let\@evenhead\@empty
 \def\@oddfoot{\centerline{\thepage}}%
 \let\@evenfoot\@oddfoot}
\makeatother

\usepackage[utf8]{inputenc}
\usepackage[T1]{fontenc}
\usepackage{amsmath}
\usepackage{booktabs}
\usepackage{dsfont}
\usepackage{graphicx}
\usepackage{color}
\usepackage{dirtytalk}
\usepackage{algorithm}
\usepackage{algorithmicx}
\usepackage[noend]{algpseudocode}
\usepackage{csquotes}
\usepackage{rotating}
\usepackage{hyperref}
\makeatletter
\providecommand{\doi}[1]{%
  \begingroup
    \let\bibinfo\@secondoftwo
    \urlstyle{rm}%
    \href{http://dx.doi.org/#1}{%
      doi:\discretionary{}{}{}%
      \nolinkurl{#1}%
    }%
  \endgroup
}
\makeatother
\usepackage{subcaption}
\usepackage{enumitem}
\usepackage{tikz}
\usetikzlibrary{patterns}
\usetikzlibrary{positioning}
\usetikzlibrary{decorations.pathreplacing,angles,quotes}
\usepackage{lineno}

\usepackage{natbib}
 \bibpunct[, ]{(}{)}{,}{a}{}{,}%

\begin{document}

\begin{frontmatter}

\title{An anytime tree search algorithm for two-dimensional two- and three-staged guillotine packing problems}

\author[]{Florian Fontan\corref{cor1}}\ead{dev@florian-fontan.fr}
\author[1]{Luc Libralesso}\ead{luc.libralesso@grenoble-inp.fr}
\cortext[cor1]{Corresponding author}
\address[1]{Univ. Grenoble Alpes, CNRS, Grenoble INP\tnoteref{t1}, G-SCOP, 38000 Grenoble, France}

\tnotetext[t1]{Institute of Engineering Univ. Grenoble Alpes}

\begin{abstract}
  \cite{libralesso_anytime_2020} proposed an anytime tree search algorithm for the 2018 ROADEF/EURO challenge glass cutting problem\footnote{\url{https://www.roadef.org/challenge/2018/en/index.php}}. The resulting program was ranked first among 64 participants. In this article, we generalize it and show that it is not only effective for the specific problem it was originally designed for, but is also very competitive and even returns state-of-the-art solutions on a large variety of Cutting and Packing problems from the literature. We adapted the algorithm for two-dimensional Bin Packing, Multiple Knapsack, and Strip Packing Problems, with two- or three-staged exact or non-exact guillotine cuts, the orientation of the first cut being imposed or not, and with or without item rotation. The combination of efficiency, ability to provide good solutions fast, simplicity and versatility makes it particularly suited for industrial applications, which require quickly developing algorithms implementing several business-specific constraints. The algorithm is implemented in a new software package called PackingSolver.
\end{abstract}
%\begin{abstract}
  %In this article, we introduce PackingSolver, a new solver for two-dimensional two- and three-staged guillotine Packing Problems. It relies on a simple yet powerful anytime tree search algorithm called Memory Bounded A* (MBA*). This algorithm was first introduced in \cite{libralesso_anytime_2020} for the 2018 ROADEF/EURO challenge glass cutting problem\footnote{\url{https://www.roadef.org/challenge/2018/en/index.php}}, for which it had been ranked first during the final phase. In this article, we generalize it for a large variety of Cutting and Packing problems. The resulting program can tackle two-dimensional Bin Packing, Multiple Knapsack, and Strip Packing Problems, with two- or three-staged exact or non-exact guillotine cuts, the orientation of the first cut being imposed or not, and with or without item rotation. Despite its simplicity and genericity, computational experiments show that this approach is competitive compared to the other dedicated algorithms from the literature. It even returns state-of-the-art solutions on several variants. The combination of efficiency, ability to provide good solutions fast, simplicity and versatility makes it particularly suited for industrial applications, which require quickly developing algorithms implementing several business-specific constraints.
%\end{abstract}

\begin{keyword}
two-dimensional guillotine packing \sep bin packing \sep knapsack \sep strip packing \sep anytime algorithm, tree search algorithm 
\end{keyword}

\end{frontmatter}

%\linenumbers

The 2018 ROADEF/EURO challenge featured an industrial glass cutting problem arising at the French company Saint Gobain. \cite{libralesso_anytime_2020} proposed an anytime tree search algorithm that was ranked first in the final phase of the challenge. They showed that the algorithm performs very well on this specific variant with the specific instances considered. Indeed, some of the industrial constraints of the problem seem to favor this kind of constructive approach. In particular, the problem includes precedence constraints, which highly penalize other approaches such as local search, dynamic programming, mixed-integer linear programming or column generation. Therefore, it was not obvious \textit{a priori} whether the algorithm would be competitive on other variants. In this article, we show that even on pure Packing Problems from the literature, it is competitive compared to the other dedicated algorithms, and is even able to return state-of-the-art solutions on several variants.

Even though most of the new constraints taken into account integrate naturally within the algorithm, several improvements need to be made to make it efficient on the large variety of problems and instances from the literature: two new guide functions are proposed to deal with instances with different item distributions; an additional guide is designed for the Knapsack objective; and some flexibility has been introduced in the symmetry breaking strategy. 

\cite{libralesso_anytime_2020} proposed an efficient algorithm for a specific problem with specific constraints and instances. Here, we propose an efficient approach which should be useful for almost any (guillotine for now) Packing Problem. Also, as discussed in Section~\ref{sec:discussion}, experimenting on all these variants greatly improved our understanding of the effectiveness of MBA* and other tree search algorithms.

\section{Introduction}
\label{sec:introduction}

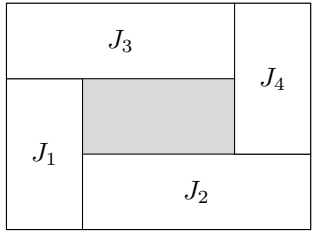
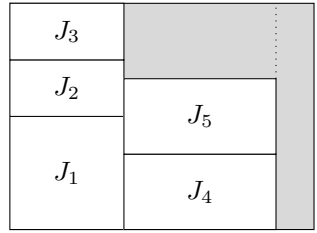
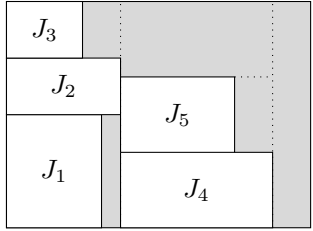
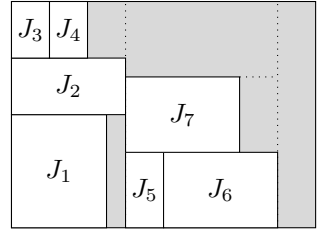
\begin{figure}
  \centering
  \begin{subfigure}[t]{0.45\textwidth}
    \centering
    \begin{tikzpicture}
      \draw[fill=gray!30] (0,0) rectangle (4,3);
      \draw[fill=white] (0,0) rectangle (1,2) node[pos=.5] {$J_1$};
      \draw[fill=white] (1,0) rectangle (4,1) node[pos=.5] {$J_2$};
      \draw[fill=white] (0,2) rectangle (3,3) node[pos=.5] {$J_3$};
      \draw[fill=white] (3,1) rectangle (4,3) node[pos=.5] {$J_4$};
    \end{tikzpicture}
    \caption{Non-guillotine pattern}
  \end{subfigure}
  $\quad$
  \begin{subfigure}[t]{0.45\textwidth}
    \centering
    \begin{tikzpicture}
      \draw[fill=gray!30] (0,0) rectangle (4,3);
      \draw[fill=white] (0,0) rectangle (1.5,1.5) node[pos=.5] {$J_1$};
      \draw[fill=white] (0,1.5) rectangle (1.5,2.25) node[pos=.5] {$J_2$};
      \draw[fill=white] (0,2.25) rectangle (1.5,3) node[pos=.5] {$J_3$};
      \draw[fill=white] (1.5,0) rectangle (3.5,1) node[pos=.5] {$J_4$};
      \draw[fill=white] (1.5,1) rectangle (3.5,2) node[pos=.5] {$J_5$};
      \draw[dotted] (3.5,2) -- (3.5,3);
    \end{tikzpicture}
    \caption{Two-staged exact guillotine pattern, first stage vertical}
  \end{subfigure}

  \bigskip

  \begin{subfigure}[t]{0.45\textwidth}
    \centering
    \begin{tikzpicture}
      \draw[fill=gray!30] (0,0) rectangle (4,3);
      \draw[fill=white] (0,0) rectangle (1.25,1.5) node[pos=.5] {$J_1$};
      \draw[fill=white] (0,1.5) rectangle (1.5,2.25) node[pos=.5] {$J_2$};
      \draw[fill=white] (0,2.25) rectangle (1,3) node[pos=.5] {$J_3$};
      \draw[fill=white] (1.5,0) rectangle (3.5,1) node[pos=.5] {$J_4$};
      \draw[fill=white] (1.5,1) rectangle (3,2) node[pos=.5] {$J_5$};
      \draw[dotted] (1.5,0) -- (1.5,3);
      \draw[dotted] (3.5,0) -- (3.5,3);
      \draw[dotted] (3,2) -- (3.5,2);
    \end{tikzpicture}
    \caption{Two-staged non-exact guillotine pattern, first stage vertical}
  \end{subfigure}
  $\quad$
  \begin{subfigure}[t]{0.45\textwidth}
    \centering
    \begin{tikzpicture}
      \draw[fill=gray!30] (0,0) rectangle (4,3);
      \draw[fill=white] (0,0) rectangle (1.25,1.5) node[pos=.5] {$J_1$};
      \draw[fill=white] (0,1.5) rectangle (1.5,2.25) node[pos=.5] {$J_2$};
      \draw[fill=white] (0,2.25) rectangle (0.5,3) node[pos=.5] {$J_3$};
      \draw[fill=white] (0.5,2.25) rectangle (1,3) node[pos=.5] {$J_4$};
      \draw[fill=white] (1.5,0) rectangle (2,1) node[pos=.5] {$J_5$};
      \draw[fill=white] (2,0) rectangle (3.5,1) node[pos=.5] {$J_6$};
      \draw[fill=white] (1.5,1) rectangle (3,2) node[pos=.5] {$J_7$};
      \draw[dotted] (1.5,0) -- (1.5,3);
      \draw[dotted] (3.5,0) -- (3.5,3);
      \draw[dotted] (3,2) -- (3.5,2);
    \end{tikzpicture}
    \caption{Three-staged exact guillotine pattern, first stage vertical}
  \end{subfigure}
  \caption{Pattern type examples}
  \label{fig::patterns}
\end{figure}

We consider two-dimensional guillotine Packing Problems: one has to pack rectangles of various sizes into larger bins while only edge-to-edge cuts are allowed.
In a solution, guillotine cuts can be partitioned into stages, \textit{i.e.}\ series of parallel cuts, and it is common to limit the number of allowed stages.
Here, we restrict to two- or three-staged guillotine patterns. In both cases, we consider both exact and non-exact variants.
In the non-exact variant, an additional cut is allowed to separate items from waste.
Figure~\ref{fig::patterns} illustrates the different pattern types.

We consider the three main objectives studied in the literature: Bin Packing, Knapsack and Strip Packing.
In Bin Packing and Strip Packing Problems, all items need to be produced. In Bin Packing Problems, the number of used bins is minimized, while in Strip Packing Problems, there is only one container with one infinite dimension and the objective is to minimize the length used in this dimension.
In Knapsack Problems, the number of containers is limited, every item has a profit and the total profit of the packed items is maximized.

Finally, for each variant, we consider the oriented case where item rotation is not allowed and non-oriented case where it is.

Throughout the article, the different variants are named following our notations illustrated with the following examples:
\begin{itemize}[noitemsep,nolistsep]
  \item BPP-O: (non-guillotine) Bin Packing Problem, Oriented
  \item G-BPP-R: Guillotine cuts, Bin Packing Problem, Rotation
  \item 2G-KP-O: 2-staged exact guillotine cuts, first cut horizontal or vertical, Knapsack Problem, Oriented
  \item 3NEGH-SPP-O: 3-staged non-exact guillotine cuts, first cut horizontal, Strip Packing Problem, Oriented
\end{itemize}

We also use the following vocabulary: a $k$-cut is a cut performed in the $k$-th stage. Cuts separate bins into $k$-th level sub-plates.
For example, $1$-cuts separate the bin in several first level sub-plates. $S$ denotes a solution or a node in the search tree (a partial solution).

\begin{figure}
  \centering
  \begin{tikzpicture}
    \draw[fill=gray!30] (0,0) rectangle (4,3);
    \draw[fill=white] (0,0) rectangle (1,3) node[pos=.5] {$J_1$};
    \draw[fill=white] (1,0) rectangle (3,1) node[pos=.5] {$J_2$};
    \draw[fill=white] (1,1) rectangle (2,2) node[pos=.5] {$J_3$};
    \draw[fill=white] (2,1) rectangle (3.5,1.5) node[pos=.5] {$J_4$};
    \draw[dotted] (3.5,0) -- (3.5,3);
    \draw[dotted] (1,2) -- (3.5,2);
    \draw (3.5,0) node[anchor=north] {$x_1^\text{curr} = x_3^\text{curr}$};
    \draw (1,0) node[anchor=north] {$x_1^\text{prev}$};
    \draw (2,0) node[anchor=north] {$x_3^\text{prev}$};
    \draw (0,1) node[anchor=east] {$y_2^\text{prev}$};
    \draw (0,2) node[anchor=east] {$y_2^\text{curr}$};
    \draw[pattern=north west lines, pattern color=blue] (0,0) rectangle (1,3);
    \draw[pattern=north west lines, pattern color=blue] (1,0) rectangle (3.5,2);
  \end{tikzpicture}
  \caption{Last bin of a solution which does not contain all items. The area is the whole hatched part and the waste in the grey hatched part.}
  \label{fig:area}
\end{figure}
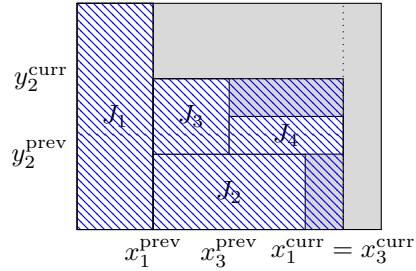

The following definitions are given for the case where the first cut in the last bin is vertical, but naturally, adapt to the case where it is horizontal.
We call the last first level sub-plate, the rightmost one containing an item; the last second level sub-plate, the topmost one containing an item in the last first level sub-plate; and the last third level sub-plate the rightmost one containing an item in the last second level sub-plate.
$x_1^\text{prev}(S)$ and $x_1^\text{curr}(S)$ are the left and right coordinates of the last first level sub-plate; $y_2^\text{prev}(S)$ and $y_2^\text{curr}(S)$ are the bottom and top coordinates of the last second level sub-plate; and $x_3^\text{prev}(S)$ and $x_3^\text{curr}(S)$ are the left and right coordinates of the last third level sub-plate. Figure \ref{fig:area} presents a usage example of these definitions. We define the area and the waste of a solution $S$ as follows:

\begin{displaymath}
    \begin{array}{lll}
        \mathrm{area}(S) &=& \left\{
            \begin{array}{llll}
              A &+& x_1^\text{curr}(S) h & \text{if $S$ contains all items} \\
              A &+& x_1^\text{prev}(S) h & \\
                &+& (x_1^\text{curr}(S) - x_1^\text{prev}(S)) y_2^\text{prev}(S) & \\
                &+& (x_3^\text{curr}(S) - x_1^\text{prev}(S)) (y_2^\text{curr}(S) - y_2^\text{prev}(S)) & \text{otherwise} \\
            \end{array}
            \right. \vspace{.3cm}\\
        \mathrm{waste}(S) &=& \mathrm{area}(S) - \mathrm{item\_area}(S)
    \end{array}
\end{displaymath}
with $A$ the sum of the areas of all but the last bin, $h$ the height of the last bin and $\mathrm{item\_area}(S)$ the sum of the area of the items of $S$. Area and waste are illustrated in Figure~\ref{fig:area}.

\section{Literature review}
\label{sec:literature}

Two-dimensional guillotine Packing Problems have been introduced by \cite{gilmore_multistage_1965} and have received a lot of attention since.
Researchers usually focus on one specific variant or only on a few ones.

Algorithms are sometimes adapted for both the oriented and the non-oriented cases. \cite{velasco_improved_2019} developed a heuristic for G-KP-O and G-KP-R,
\cite{wei_block-based_2014} for G-SPP-O and G-SPP-R, \cite{charalambous_constructive_2011}, \cite{fleszar_three_2013} and
\cite{cui_applying_2018} for G-BPP-O and G-BPP-R, \cite{lodi_integer_2003} an exact algorithm for 2NEGH-KP-O and 2NEGH-KP-R.

Some methods have been designed to work on more variants. \cite{do_nascimento_mip-cp_2019} developed an exact algorithm for G-KP-O, 3NEGH-KP-O, 2NEGH-KP-O and the three-dimensional variants, \cite{bortfeldt_genetic_2009} developed a genetic algorithm for G-KP-O, G-KP-R, and the non-guillotine variants. \cite{alvelos_sequence_2009} and \cite{silva_integer_2010} respectively developed a heuristic and an exact algorithm for 3NEGH-BPP-O, 3GH-BPP-O, 2NEGH-BPP-O and 2GH-BPP-O, and the non-oriented cases. \cite{furini_modeling_2016} introduced a model for G-KP-O and G-SPP-O. \cite{lodi_models_2004} proposed a unified tabu search for two- and three-dimensional Packing Problems. They provide computational experiments for BPP-O and the three-dimensional variant. They also describe how to adapt the algorithm for several variants such as Strip Packing or Multiple Knapsack. However, adapting the algorithm requires to provide a heuristic procedure, on which the efficiency of the algorithm highly relies. We did not find any use of their tabu search in the subsequent literature. Also, a framework has been proposed by \cite{nepomuceno_hybrid_2008}; unfortunately, it has only been implemented for BPP-O and we did not find any use of their framework in the subsequent literature either.

\bigskip

Regarding our methodology, even though tree search algorithms have been widely used to solve Packing Problems, the search algorithm that we implemented does not seem to have been proposed before.
We may notice that many packing algorithms rely on Beam Search which is relatively close, as discussed in Section~\ref{sec:discussion}. \cite{akeb_beam_2009}, \cite{hifi_beam_2009}, \cite{akeb_adaptive_2010} and \cite{akeb_augmented_2011} implemented it for Circular Packing Problems; \cite{bennell_beam_2010} and \cite{bennell_beam_2018} for Irregular Packing Problems; \cite{wang_multi-round_2013}, \cite{araya_beam_2014} and \cite{araya_beam_2020} for three-dimensional Packing Problems; and \cite{hifi_parallel_2012} for 2NEGH-KP-O.
However, these Beam Search implementations significantly differ from our tree search implementation. Most of them do not use a restart strategy, are block-based approaches and use probing (filling partial solutions with a greedy heuristic) to evaluate the quality of nodes. Furthermore, they are globally more complex than our tree search implementation, suggesting that we better captured the key ideas that make tree search algorithms efficient for Packing Problems.

\section{Algorithm description}
\label{sec:algorithm}

We propose an anytime tree search algorithm.

Anytime is a terminology usually found in automated planning and scheduling (AI planning) communities.
It means that the algorithm can be stopped at any time and still provides good solutions.
In other words, it produces feasible solutions quickly and improves them over time (as classical meta-heuristics do).

Tree search algorithms represent the solution space as an implicit tree called \say{branching scheme} and explore it completely in the case of exact methods
or partially in the case of heuristic methods. The branching scheme is described in Section~\ref{ssec:branching_scheme} and the tree search algorithm in Section~\ref{ssec:tree_search}.

\subsection{Branching scheme}
\label{ssec:branching_scheme}

We describe the branching scheme for the 3-staged cases with vertical cuts in the first stage. For the 2-staged cases, we merely impose the position of the first cut to be at the end of the bin and adjust the computation of parameters accordingly; and when the cuts in the first stage are horizontal, we simply adapt the computation of coordinates.

The branching scheme is rather straightforward. The root node is the empty solution without any items, and at each stage, a new item is added.
All items that do not belong to the current node are considered.
However, items in a solution are inserted according to the following order: rightmost first level sub-plates first; within a first level sub-plate, bottommost second level sub-plates first; and within a second level sub-plate, rightmost items first. Thus, a new item can be inserted in a new bin; in a new first level sub-plate to the right of the current one; in a new second-level sub-plate above the current one; in a new third-level sub-plate, to the right of the last added item. If the cuts of the first stage can be vertical or horizontal, then two different insertions in a new bin are considered: an insertion in a new bin with vertical cuts in the first stage, and an insertion in a new bin with horizontal cuts in the first stage.

To handle exact guillotine cuts, we simply fix the position of the 2-cut above an item inserted in a new bin, first or second level sub-plate,
\textit{i.e.}\ the next items inserted in the same second level sub-plate will only be those of the same height.

Item rotation or not is naturally handled in the branching scheme.

\bigskip

To reduce the size of the tree, we apply some simple dominance rules.

First, if an item can be inserted in the current bin, we do not consider insertions in a new bin; and if an item can be inserted in the current first (resp.\ second) level sub-plate without increasing the position of its left $1$-cut (resp.\ top $2$-cut), we do not consider insertions in a new first (resp.\ second) level sub-plate.

\begin{figure}
  \centering
  \begin{subfigure}[t]{0.45\textwidth}
    \centering
    \begin{tikzpicture}
      \draw[fill=gray!30] (0,0) rectangle (4,3);
      \draw[fill=white] (0,0) rectangle (1,3) node[pos=.5] {$J_1$};
      \draw[fill=white] (1,0) rectangle (3,1) node[pos=.5] {$J_2$};
      \draw[fill=white] (1,1) rectangle (2,2) node[pos=.5] {$J_3$};
      \draw[fill=white] (2,1) rectangle (2.5,2) node[pos=.5] {$J_4$};
      \draw[dotted] (3,1) -- (3,3);
      \draw[dotted] (2.5,2) -- (3,2);
    \end{tikzpicture}
    \caption{}
  \end{subfigure}
  \begin{subfigure}[t]{0.45\textwidth}
    \centering
    \begin{tikzpicture}
      \draw[fill=gray!30] (0,0) rectangle (4,3);
      \draw[fill=white] (0,0) rectangle (1,3) node[pos=.5] {$J_1$};
      \draw[fill=white] (1,0) rectangle (3,1) node[pos=.5] {$J_2$};
      \draw[fill=white] (1,1) rectangle (2,2) node[pos=.5] {$J_3$};
      \draw[fill=white] (2,1) rectangle (3,1.5) node[pos=.5] {$J_4$};
      \draw[dotted] (3,1.5) -- (3,3);
      \draw[dotted] (2,2) -- (3,2);
      \draw[pattern=north west lines, pattern color=blue] (2,1.5) rectangle (3,2);
    \end{tikzpicture}
    \caption{}
  \end{subfigure}
  \caption{Solution (a) dominates solution (b) because the hatched area will not be used}
  \label{fig:dominance}
\end{figure}
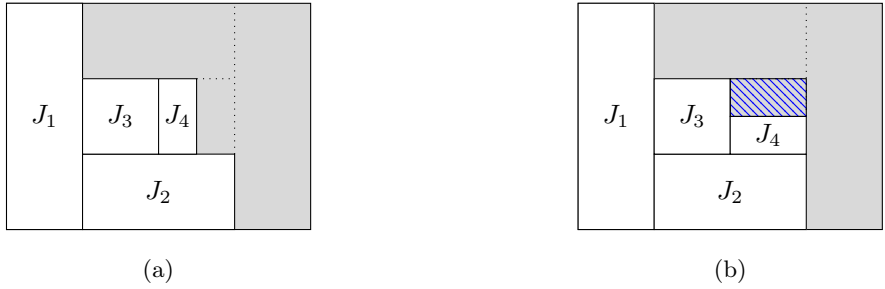

Then, if item rotation is allowed, some insertions can be discarded as illustrated in Figure~\ref{fig:dominance}.

We also impose an order on identical items.

Finally, we add the following symmetry breaking strategy: a $k$-level sub-plate is forbidden to contain an item with a smaller index than the previous $k$ level sub-plate of the same $(k-1)$-level sub-plate.
The symmetry breaking strategy is controlled with a parameter $s$, $1 \le s \le 4$. If $s = k$, then the symmetry breaking strategy is only used with $k'$ level sub-plates, $k' \ge k$. For example, if $s = 4$, no symmetry breaking strategy is used. The choice of the value of $s$ is discussed in Section~\ref{sec:discussion}.

\subsection{Tree search algorithm}
\label{ssec:tree_search}

The tree described in the previous section is too large to be entirely explored. Therefore, we use a tree search algorithm that we called Memory Bounded A* (MBA*) to explore the most interesting parts in priority. The pseudo-code is given in Algorithm~\ref{MBA}. MBA* starts with a queue containing only the root node. At each iteration, the \say{best} node is extracted from the queue and its children are added to the queue.
If the size of the queue goes over a pre-defined threshold value, the \say{worst} nodes are discarded. We start with a threshold of 2, and each time the queue becomes empty, we start over with a threshold multiplied by the growth factor $f$. We choose $f = 1.5$ as discussed in Section~\ref{sec:discussion}.

\begin{algorithm}[ht]
  \caption{Memory Bounded A* (MBA*)}
  \label{MBA}
  \begin{algorithmic}[1]
    \State $\mathrm{queue} \gets \left\{ \mathrm{root} \right\}$
    \While{$|\mathrm{queue}| \neq \emptyset$ and $\mathrm{time} < \mathrm{timelimit}$}
    \State $n \gets \mathrm{extractBest}(\mathrm{queue})$
    \State $\mathrm{queue} \gets \mathrm{queue} \setminus \{ n \}$
    \ForAll{ $v \in children(n)$ }
    \State $\mathrm{queue} \gets \mathrm{queue} \cup \left\{ v \right\}$
    \EndFor
    \While { $|\mathrm{queue}| > D$ }
    \State $n \gets \mathrm{extractWorst}($queue$)$
    \State $\mathrm{queue} \gets \mathrm{queue} \setminus \left\{ n \right\}$
    \EndWhile
    \EndWhile
  \end{algorithmic}
\end{algorithm}

The function used to define \say{better} and \say{worse} is called a guide. The lower the value of the guide function is, the better the solution.
For Bin Packing and Strip Packing Problems, we designed the following guide functions:
\begin{displaymath}
  c_0(S) = \mathrm{waste\_percentage}(S)
\end{displaymath}
\begin{displaymath}
  c_1(S) = \frac{\mathrm{waste\_percentage}(S)}{\mathrm{mean\_item\_area}(S)}
\end{displaymath}
\begin{displaymath}
  c_2(S) = \frac{\mathrm{0.1 + waste\_percentage}(S)}{\mathrm{mean\_item\_area}(S)}
\end{displaymath}
\begin{displaymath}
  c_3(S) = \frac{\mathrm{0.1 + waste\_percentage}(S)}{\mathrm{mean\_squared\_item\_area}(S)}
\end{displaymath}
with
\begin{itemize}[noitemsep,nolistsep]
  \item $\mathrm{waste\_percentage}(S) = \mathrm{waste}(S) / \mathrm{area}(S)$;
  \item $\mathrm{mean\_item\_area}(S)$ the mean area of the items of $S$;
  \item $\mathrm{mean\_squared\_item\_area}(S)$ the mean squared area of the items of $S$.
\end{itemize}

For Knapsack Problems, we use the following guide function:
\begin{displaymath}
  c_4(S) = \frac{\mathrm{area}(S)}{\mathrm{profit}(S)}
\end{displaymath}
with $\mathrm{profit}(S)$ the sum of profit of the items of $S$.

The importance and design of these guide functions are discussed in Section~\ref{sec:discussion}.

\section{Computational experiments}
\label{sec:experiments}

The algorithm has been implemented in C++ in a new software package called PackingSolver.
The code is available online\footnote{\url{https://github.com/fontanf/packingsolver}}.
The repository also contains all the scripts used to conduct the experiments so that results can be reproduced. The results presented above have been obtained with PackingSolver $0.2$\footnote{\url{https://github.com/fontanf/packingsolver/releases/tag/0.2}} running on a personal computer with an Intel Core i5-8500 CPU @ 3.00GHz $\times$ 6.
We allow running up to 3 threads with different settings in parallel. The settings have been chosen following the observations given in Section~\ref{sec:discussion}. Better settings may exist, we try to reproduce the results one would obtain in a practical situation where the global characteristics of the instances are known.

We compare the performances of our algorithm with the best algorithms from the literature for each variant. Due to a large number of problems, we only provide a synthesis of the results here. However, detailed results are available online\footnote{\url{https://github.com/fontanf/packingsolver/blob/0.2/results_rectangleguillotine.ods}} and the interested reader is encouraged to have a look at them.

Results are summarized in Tables~\ref{tab:bpp}, \ref{tab:kp} and \ref{tab:spp}.
The first column of the tables indicates the article from which the results have been extracted or the parameters we used for our algorithm. $c_a^b$ indicates a thread with guide function $c_a$ and symmetry breaking parameter $b$.  TL stands for \say{time limit}. The time limit has been chosen to yield a good compromise between computation time and the best solution value. 
We only indicate the frequencies of the processors used to evaluate the other algorithms when they significantly differ from ours, \textit{i.e.}\ below $2$GHz.

For Bin Packing Problems, the second column contains the total number of bins used in Table~\ref{tab:bpp_a} and the average of
the average percentage of waste of each sub-dataset in Table~\ref{tab:bpp_b}. For Knapsack and Strip Packing Problems, it contains
the average gap to the best-known solutions. The third one indicates the average time to best when available, or the average computation time.

Dataset \say{hifi} is a dataset composed of instances from
\cite{christofides_algorithm_1977},
\cite{wang_two_1983},
\cite{oliveira_improved_1990},
\cite{tschoke_new_1995},
\cite{fekete_new_1997},
\cite{fayard_efficient_1998},
\cite{hifi_improvement_1997} and
\cite{cung_constrained_2000}. Researchers usually test their algorithms on a subset of these instances, but often not the same.
Dataset \say{bwmv} refers to datasets from \cite{berkey_two-dimensional_1987} and \cite{martello_exact_1998} which are usually used together.

Other datasets are
\begin{itemize}[noitemsep,nolistsep]
  \item \say{beasley1985} from \cite{beasley_algorithms_1985}
  \item \say{fayard1998} from \cite{fayard_efficient_1998}
  \item \say{kroger1995} from \cite{kroger_guillotineable_1995}
  \item \say{hopper2000} from \cite{hopper_two-dimensional_2000}
  \item \say{hopper2001} from \cite{hopper_empirical_2001}
  \item \say{alvarez2002} from \cite{alvarez-valdes_tabu_2002}
  %\say{gcutd} from \cite{cintra_algorithms_2008},
  \item \say{morabito2010} from \cite{morabito_heuristic_2010}
  \item \say{hifi2012} from \cite{hifi_parallel_2012}
  \item \say{velasco2019} from \cite{velasco_improved_2019}
\end{itemize}

\subsection{Bin Packing Problems}

Results for Bin Packing Problems are summarized in Table~\ref{tab:bpp}.
On 2NEGH-BPP-O and 2NEGH-BPP-R, PackingSolver respectively needs fewer bins than the algorithms from \cite{cui_heuristic_2013-1} and \cite{cui_hybrid_2016} for the considered datasets. Furthermore, the average time to best is of the order of a second, which is significantly smaller than the average time reported for the other algorithms.
On 3NEGH-BPP-O, 3GH-BPP-O, and 2NEGH-BPP-O, the average of the average percentage of waste of PackingSolver is smaller than the one of the algorithms from \cite{alvelos_sequence_2009}. However, on 2GH-BPP-O, it is greater.
Finally, compared to the algorithms from \cite{puchinger_models_2007} and \cite{alvelos_hybrid_2014}, it needs more bins, but the average time to best is two orders of magnitude smaller than the average time reported for those algorithms. We also note that PackingSolver respectively needs significantly fewer bins on 3NEGH-BPP-O and 3GH-BPP-O compared to the algorithms from \cite{puchinger_models_2007} and \cite{alvelos_hybrid_2014} for 3GH-BPP-O and 2NEGH-BPP-O, 
\begin{table}
  \centering
  \small
  \begin{subfigure}[t]{0.45\textwidth}
    \centering
    \begin{tabular}{l|l|l}
      \toprule
      Article / Parameters & Total & Time (s) \\
      \midrule
      %\multicolumn{3}{c}{G-BPP-O, \say{bwmv}} \\
      %\cite{hong_hybrid_2014} & $7272$ & \\
      %\cite{lodi_partial_2017} (TL $60$s) & $7289$ & \\
      %\cite{lodi_partial_2017} (TL $600$s) & $7286$ & \\
      %\cite{lodi_partial_2017} (TL $1800$s) & $7281$ & \\
      %\cite{cui_applying_2018} & $7238$ & $0.85$ \\
      %3NEGH-BPP-O, $c_0^2c_2^2c_3^3$, TL $60s$ & $7274$ & $1.366$ \\
      %\midrule
      %\multicolumn{3}{c}{G-BPP-R, \say{bwmv}} \\
      %\cite{cui_sequential_2015} (TL $50$s) & $7035$ & $20.72$ \\
      %\cite{cui_applying_2018} & $6989$ & $1.34$ \\
      %3NEGH-BPP-O, $c_0^2c_2^2c_3^3$, TL $60s$ & $7015$ & $0.660$ \\
      %\midrule
      \multicolumn{3}{c}{3NEGH-BPP-O, \say{bwmv}} \\
      PS, $c_0^2c_2^2c_3^3$, TL $60s$ & $7278$ & $0.790$ \\
      \midrule
      \multicolumn{3}{c}{3GH-BPP-O, \say{bwmv}} \\
      \cite{puchinger_models_2007} & $7325$ & $160.68$ \\
      PS, $c_0^2c_2^2c_3^3$, TL $60s$ & $7344$ & $0.808$ \\
      \midrule
      \multicolumn{3}{c}{2NEGH-BPP-O, \say{bwmv}} \\
      \cite{alvelos_hybrid_2014} & $7372$ & $29.42$ \\
      \cite{alvelos_hybrid_2014} & $7364$ & $84.04$ \\
      PS, $c_0^2c_2^2c_3^3$, TL $60s$ & $7391$ & $0.814$ \\
      \multicolumn{3}{c}{2NEGH-BPP-O, \say{hifi}} \\
      \cite{cui_heuristic_2013-1} & $260$ & $0.19$ \\
      PS, $c_2^3c_3^3c_3^4$, TL $10s$ & $255$ & $0.106$ \\
      \multicolumn{3}{c}{2NEGH-BPP-O, \say{alvarez2002}} \\
      \cite{cui_heuristic_2013-1} & $219$ & $9.5$ \\
      PS, $c_2^3c_3^3c_3^4$, TL $10s$ & $218$ & $0.346$ \\
      \midrule
      \multicolumn{3}{c}{2NEGH-BPP-R, \say{bwmv}} \\
      \cite{cui_hybrid_2016} & $7034$ & $20.72$ \\
      PS, $c_0^2c_2^2c_3^3$, TL $60s$ & $7029$ & $0.590$ \\
      \bottomrule
    \end{tabular}
    \caption{}
    \label{tab:bpp_a}
  \end{subfigure}
  \begin{subfigure}[t]{0.45\textwidth}
    \centering
    \begin{tabular}{l|l|l}
      \toprule
      Article / Parameters & Waste & Time (s) \\
      \midrule
      \multicolumn{3}{c}{3NEGH-BPP-O, \say{bwmv}} \\
      \cite{alvelos_sequence_2009} & $26.52$ & \\
      PS, $c_0^2c_2^2c_3^3$, TL $60s$ & $20.93$ & $0.790$ \\
      \midrule
      \multicolumn{3}{c}{3GH-BPP-O, \say{bwmv}} \\
      \cite{alvelos_sequence_2009} & $26.29$ & \\
      PS, $c_0^2c_2^2c_3^3$, TL $60s$ & $22.34$ & $0.808$ \\
      \midrule
      \multicolumn{3}{c}{2NEGH-BPP-O, \say{bwmv}} \\
      \cite{alvelos_sequence_2009} & $26.12$ & \\
      PS, $c_0^2c_2^2c_3^3$, TL $60s$ & $23.21$ & $0.807$ \\
      \midrule
      \multicolumn{3}{c}{2GH-BPP-O, \say{bwmv}} \\
      \cite{alvelos_sequence_2009} & $49.06$ & \\
      PS, $c_0^3c_2^3c_3^4$, TL $60s$ & $49.45$ & $0.181$ \\
      \bottomrule
    \end{tabular}
    \caption{}
    \label{tab:bpp_b}
  \end{subfigure}
  \caption{Results on Bin Packing Problems}
  \label{tab:bpp}
\end{table}

\subsection{Knapsack Problems}

Results for Knapsack Problems are summarized in Table~\ref{tab:kp}.
We include comparisons with algorithms designed for the non-staged variants.
In these cases, PackingSolver usually fails to find the best solutions. It seems likely that they often cannot be reached with only 3 stages. However, its average gap to best is generally less than $1\%$ and on datasets \say{velasco2019} it is even better than the recent algorithm from \cite{velasco_improved_2019}.
The same happens on dataset \say{fayard1998} for G-KP-R, but the algorithm developed by \cite{bortfeldt_genetic_2009} seems to perform significantly worse than more recent algorithms and none of them has been tested on this dataset.

On 3NEGV-KP-O, the average gap to best of PackingSolver is better than \cite{cui_heuristic_2015}, but at the expense of longer computation times.
For 2NEGH-KP-O, as \cite{alvarez-valdes_grasp_2007}, it finds all the best solutions, but faster. Compared to the algorithm from \cite{hifi_algorithms_2008}, it performs slightly worse on dataset \say{alvarez2002} (even if the average gap is $0.0$, it fails to find the best solution on two instances) but better on dataset \say{hifi2012}.

On variants 2NEG-KP-R, 2G-KP-O, 2GH-KP-O, and 2GV-KP-O for which \cite{lodi_integer_2003} and \cite{hifi_approximate_2001} developed exact algorithms, PackingSolver finds all optimal solutions in reasonable computation times.

\begin{table}
  %\SingleSpacedXI
  \centering
  \small
  \begin{tabular}{l|l|l}
    \toprule
    Article / Parameters & Gap & Time (s) \\
    \midrule
    \multicolumn{3}{c}{G-KP-O, \say{fayard1998}} \\
    \cite{velasco_improved_2019} & $0.00$ & $0.06$ \\
    PS, 3NEG-KP-O, $c_4^2c_4^3$, TL $10$s & $0.16$ & $0.182$ \\
    \multicolumn{3}{c}{G-KP-O, \say{alvarez2002}} \\
    \cite{wei_bidirectional_2015} & $0.02$ & $21.987$ \\
    \cite{velasco_improved_2019} & $0.00$ & $93.681$ \\
    PS, 3NEG-KP-O, $c_4^2c_4^3$, TL $60$s & $0.48$ & $13.264$ \\
    \multicolumn{3}{c}{G-KP-O, \say{hopper2001}} \\
    \cite{wei_bidirectional_2015} & $0.31$ & $22.214$ \\
    PS, 3NEG-KP-O, $c_4^2c_4^3$, TL $10$s & $4.69$ & $1.283$ \\
    \multicolumn{3}{c}{G-KP-O, \say{morabito2010}} \\
    \cite{velasco_improved_2019} & $0.01$ & $19.57$ \\
    PS, 3NEG-KP-O, $c_4^2c_4^3$, TL $10$s & $0.17$ & $0.332$ \\
    \multicolumn{3}{c}{G-KP-O, \say{beasley1985}} \\
    \cite{dolatabadi_exact_2012} & $0.00$ & $1397.738$ \\
    \cite{wei_bidirectional_2015} & $0.44$ & $20.923$ \\
    PS, 3NEG-KP-O, $c_4^2c_4^3$, TL $10$s & $0.56$ & $0.204$ \\
    \multicolumn{3}{c}{G-KP-O, \say{velasco2019}} \\
    \cite{velasco_improved_2019} & $1.42$ & $165.618$ \\
    PS, 3NEG-KP-O, $c_4^2c_4^3$, TL $120$s & $0.47$ & $34.682$ \\
    \midrule
    \multicolumn{3}{c}{G-KP-R, \say{hopper2001}} \\
    \cite{wei_bidirectional_2015} & $0.00$ & $5.04$ \\
    PS, 3NEG-KP-R, $c_4^2c_4^3$, TL $30$s & $1.71$ & $8.049$ \\
    \multicolumn{3}{c}{G-KP-R, \say{fayard1998}} \\
    \cite{bortfeldt_genetic_2009} & $1.57$ &  \\
    PS, 3NEG-KP-R, $c_4^2c_4^3$, TL $30$s & $0.00$ & $2.578$ \\
    \multicolumn{3}{c}{G-KP-R, \say{velasco2019}} \\
    \cite{velasco_improved_2019} & $1.05$ & $170.20$ \\
    PS, 3NEG-KP-R, $c_4^2c_4^3$, TL $120$s & $0.51$ & $38.590$ \\
    \midrule
    \multicolumn{3}{c}{3NEGV-KP-O, \say{alvarez2002}} \\
    \cite{cui_heuristic_2015} & $0.09$ & $2.06$ \\
    PS, $c_4^1c_4^2c_4^3$, TL $60$s & $0.01$ & $11.879$ \\
    \bottomrule
  \end{tabular}
  \begin{tabular}{l|l|l}
    \toprule
    Article / Parameters & Gap & Time (s) \\
    \midrule
    \multicolumn{3}{c}{2NEGH-KP-O, \say{hifi}} \\
    \cite{alvarez-valdes_grasp_2007} & $0.00$ & $0.5$ \\
    PS, $c_4^2c_4^3$, TL $3$s & $0.00$ & $0.032$ \\
    \multicolumn{3}{c}{2NEGH-KP-O, \say{alvarez2002}} \\
    \cite{hifi_algorithms_2008} & $0.00$ & $0.2$ \\
    PS, $c_4^2c_4^3$, TL $10$s & $0.00$ & $0.410$ \\
    \multicolumn{3}{c}{2NEGV-KP-O, \say{alvarez2002}} \\
    \cite{hifi_algorithms_2008} & $0.00$ & $0.2$ \\
    PS, $c_4^2c_4^3$, TL $10$s & $0.00$ & $0.382$ \\
    \multicolumn{3}{c}{2NEGH-KP-O, \say{hifi2012}} \\
    \cite{hifi_algorithms_2008} & $0.26$ & $368.365$ \\
    PS, $c_4^2c_4^3$, TL $300$s & $0.12$ & $138.742$ \\
    \multicolumn{3}{c}{2NEGV-KP-O, \say{hifi2012}} \\
    \cite{hifi_algorithms_2008} & $0.24$ & $310.105$ \\
    PS, $c_4^2c_4^3$, TL $300$s & $0.00$ & $121.014$ \\
    \midrule
    \multicolumn{3}{c}{2NEGH-KP-R, \say{hifi}} \\
    \cite{lodi_integer_2003} (533 MHz) & $0.00$ & $34.348$ \\
    PS, $c_4^2c_4^3$, TL $3$s & $0.00$ & $0.161$ \\
    \midrule
    \multicolumn{3}{c}{2G-KP-O, \say{hifi}} \\
    \cite{hifi_approximate_2001} (250 Mhz) & $0.00$ & $1.253$ \\
    PS, $c_4^2c_4^3$, TL $1$s & $0.00$ & $0.003$ \\
    \midrule
    \multicolumn{3}{c}{2GH-KP-O, \say{hifi}} \\
    \cite{hifi_approximate_2001} (250 Mhz) & $0.00$ & $1.145$ \\
    PS, $c_4^2c_4^3$, TL $1$s & $0.00$ & $0.002$ \\
    \multicolumn{3}{c}{2GV-KP-O, \say{hifi}} \\
    \cite{hifi_approximate_2001} (250 Mhz) & $0.00$ & $1.147$ \\
    PS, $c_4^2c_4^3$, TL $1$s & $0.00$ & $0.005$ \\
    \bottomrule
  \end{tabular}
  \caption{Results on Knapsack Problems}
  \label{tab:kp}
  %\DoubleSpacedXI
\end{table}

Note that, to the best of our knowledge, only \cite{cui_algorithm_2008} proposed an algorithm for a variant of a Multiple Knapsack Problem. However, they consider homogenous T-shaped patterns which we do not consider in this article.

\subsection{Strip Packing Problems}

Not many variants of guillotine Strip Packing Problems have been studied in the literature; only G-SPP-O, G-SPP-R, and 2NEGH-SPP-O. This makes comparisons with PackingSolver difficult since it is limited to three-staged patterns, and 2NEGH-SPP-O has several specific structural properties that dedicated algorithms can exploit, but not a more generic one. We still provide computational experiments for these variants in Table~\ref{tab:spp}. As expected, PackingSolver does not perform as well. Still, on dataset \say{bwmv}, it returns strictly better average solutions on 16 out of 50 groups of instances for G-SPP-O and on 14 out of 50 groups of instances for G-SPP-R than the algorithm from \cite{wei_block-based_2014}. To highlight a bit more the contribution of our algorithm for Strip Packing Problems, we provide a comparison of the solutions from \cite{lodi_models_2004} and from \cite{cui_triple-solution_2017} for 2NEGH-SPP-O with the solutions returned by PackingSolver for 2NEGH-SPP-R, \textit{i.e.}\ when item rotation is allowed. The average solutions returned by PackingSolver are strictly better on each of the 50 groups of instances of dataset \say{bwmv}.

\begin{table}
  \centering
  \small
  \begin{tabular}{l|l|l}
    \toprule
    Article / Parameters & Gap & Time (s) \\
    \midrule
    \multicolumn{3}{c}{G-SPP-O, \say{kroger1995}} \\
    \cite{wei_block-based_2014} & $0.27$ & $22.67$ \\
    PS, 3NEGH-SPP-O, $c_0^2c_0^3c_0^4$, TL $30$s & $3.65$ & $10.416$ \\
    \multicolumn{3}{c}{G-SPP-O, \say{hopper2001}} \\
    \cite{wei_block-based_2014} & $0.00$ & $6.267$ \\
    PS, 3NEGH-SPP-O, $c_0^2c_0^3c_0^4$, TL $30$s & $6.75$ & $4.364$ \\
    \multicolumn{3}{c}{G-SPP-O, \say{hopper2000}} \\
    \cite{wei_block-based_2014} & $0.00$ & $20.647$ \\
    PS, 3NEGH-SPP-O, $c_0^2c_0^3c_0^4$, TL $30$s & $8.72$ & $5.899$ \\
    \multicolumn{3}{c}{G-SPP-O, \say{bwmv}} \\
    \cite{wei_block-based_2014} & $0.15$ & $17.736$ \\
    PS, 3NEGH-SPP-O, $c_0^2c_5^2c_6^3$, TL $60$s & $1.10$ & $12.831$ \\
    \midrule
    \multicolumn{3}{c}{G-SPP-R, \say{kroger1995}} \\
    \cite{cui_heuristic_2013} & $0.00$ & $56$ \\
    PS, 3NEGH-SPP-R, $c_0^2c_0^3c_0^4$, TL $30$s & $1.84$ & $9.716$ \\
    \multicolumn{3}{c}{G-SPP-R, \say{hopper2001}} \\
    \cite{wei_block-based_2014} & $0.00$ & $13.466$ \\
    3NEGH-SPP-R, $c_0^2c_0^3c_0^4$, TL $30$s & $3.00$ & $4.153$ \\
    \multicolumn{3}{c}{G-SPP-R, \say{hopper2000}} \\
    \cite{wei_block-based_2014} & $0.00$ & $13.465$ \\
    PS, 3NEGH-SPP-R, $c_0^2c_0^3c_0^4$, TL $30$s & $3.30$ & $10.7$ \\
    \multicolumn{3}{c}{G-SPP-R, \say{bwmv}} \\
    \cite{wei_block-based_2014} & $0.13$ & $18.253$ \\
    PS, 3NEGH-SPP-R, $c_0^2c_5^2c_6^3$, TL $30$s & $0.58$ & $12.592$ \\
    \bottomrule
  \end{tabular}
  \begin{tabular}{l|l|l}
    \toprule
    Article / Parameters & Gap & Time (s) \\
    \midrule
    \multicolumn{3}{c}{2NEGH-SPP-O, \say{alvarez2002}} \\
    \cite{cui_heuristic_2013} & $0.02$ & $4.78$ \\
    PS, $c_4^1c_4^2c_4^3$, TL $30$s & $1.13$ & $3.726$ \\
    \multicolumn{3}{c}{2NEGH-SPP-O, \say{bwmv}} \\
    \cite{lodi_models_2004} & $0.02$ & $66.71$ \\
    \cite{cui_triple-solution_2017} & $0.13$ & $1.77$ \\
    PS, $c_0^2c_5^2c_6^3$, TL $30$s & $0.68$ & $0.992$ \\
    \midrule
    \multicolumn{3}{c}{2NEGH-SPP-R, \say{bwmv}} \\
    \cite{lodi_models_2004} & $7.96$ & $66.71$ \\
    \cite{cui_triple-solution_2017} & $8.08$ & $1.77$ \\
    PS, $c_0^2c_5^2c_6^3$, TL $30$s & $0.00$ & $1.773$ \\
    \bottomrule
  \end{tabular}
  \caption{Results on Strip Packing Problems}
  \label{tab:spp}
\end{table}

\section{Discussion}
\label{sec:discussion}

In this section, we discuss some items related to the algorithm.

\smallskip

\paragraph{Growth factor of the queue size threshold:} In Section~\ref{ssec:branching_scheme}, we indicated that we set the growth factor of the queue size threshold to $1.5$. The greater the threshold, the better the solutions will be, but the longer MBA* will take to terminate. Furthermore, for Bin Packing and Strip Packing Problems, full solutions are usually found shortly before it terminates. Therefore, by choosing a too large value for the growth factor, we take the risk to reach the time limit having to spend a lot of time with a given threshold without obtaining any solutions from it. On the other hand, if the growth factor is too small, then only small thresholds value will be explored and no good solutions will be found. In our experiments, 1.5 proved to be a good compromise. 

\smallskip

\paragraph{Choice of guide functions:} The effectiveness of MBA* highly relies on the definition of its guide function. For MBA*, the guide function should be relevant to compare two nodes at different stages of the tree. Therefore, the waste-percentage $c_0$ appears much more relevant than the waste alone for Bin Packing and Strip Packing variants. Guide function $c_1$ is adapted from $c_0$, but it favours solutions containing larger items. This helps to avoid situations where all small items are packed in the first bins and the last bins get all the large items, creating large waste areas. Guide function $c_2$ is adapted from $c_1$: indeed, even if $c_1$ favors large items first, solutions with no waste at all will always be extracted first, even if they contain only small items. The constant in $c_2$ aims at fixing this behavior and will lead to better solutions on instances in which optimal solutions contain significant waste (more than $10\%$). $c_3$ is adapted from $c_2$ and favours even more large items first. This guide function is useful for some instances containing several very large items.
Finally, $c_4$ is a natural adaption of $c_0$ for Knapsack variants.
An experimental comparison of several guide functions for the 2018 ROADEF/EURO challenge glass cutting problem is presented in \cite{libralesso_anytime_2020}.

\smallskip

\paragraph{Depth of the symmetry breaking strategy:} In exact tree search algorithms, it is usually worth breaking symmetries. However, this is not the case when the tree is not meant to be explored completely. For example, consider two symmetrical nodes, the first one normally appearing in the queue, but the second one never being added to the queue because one of its ancestors has been removed to reduce the size of the queue. If the first one is not explored because the symmetry has been detected, then this solution will not be found during the search. How to determine the ideal depth of the symmetry breaking strategy for an instance is not clear yet. The relative size of the items compared to the bin might be an influential factor. For the experiments, we chose $2$ or $3$ as \say{standard} values. For some instances containing many items (more than $1000$), only a value of $4$ ensures finding a feasible solution quickly; in contrast, for some knapsack instances with few first-level sub-plates, a value of $1$ gives access to better solutions.
An experimental evaluation of the influence of the symmetry breaking strategy for the 2018 ROADEF/EURO challenge glass cutting problem is presented in \cite{libralesso_anytime_2020}.

\smallskip

\paragraph{MBA* vs Beam Search:} Beam Search is another popular tree search algorithm in the packing literature. Beam Search also starts with a queue containing only the root node. However, at each iteration, all nodes of the queue are expanded, and as in MBA*, if the size of the queue goes over a pre-defined threshold, the worst nodes are discarded. Thus, at each iteration, the queue always contains nodes belonging to the same level of the tree. Beam Search seems therefore effective when the guide function is relevant to compare nodes belonging to the same level. This is for example generally not the case in Branch-and-Cut trees where branching consists in fixing a variable to 0 or 1. With our branching scheme for Packing Problems, it is easier to compare such solutions, but the guide functions we presented in Section~\ref{sec:algorithm} make it even possible to compare nodes at different levels of the search tree. Thus, Beam Search expands many nodes which are not that much interesting, whereas MBA* always expands only the best current node. An experimental comparison of MBA* and Beam Search for the 2018 ROADEF/EURO challenge glass cutting problem is presented in \cite{libralesso_anytime_2020}. It shows that MBA* finds significantly better solutions than the equivalent Beam Search implementation.

\smallskip

\paragraph{Higher staged guillotine cuts:} Our branching scheme generates up to three-staged patterns. One could wonder whether it could be possible to adapt it for four-staged or non-staged guillotine patterns. However, if a similar branching scheme seems possible, it may significantly increase symmetry issues. We believe that this would be prohibitive. MBA* might be used to solve these variants, but new branching schemes need to be designed.

\smallskip

\paragraph{Item-based vs block-based:} Many researchers highlighted the benefits of using block-based approaches, \textit{i.e.}\ inserting several items at each stage of the tree \citep{bortfeldt_tree_2012,wei_block-based_2014,lodi_partial_2017}. It is interesting to note that it is not what we implemented, yet our algorithm is competitive.

\section{Conclusion and future work}

We showed that the algorithm proposed by \cite{libralesso_anytime_2020} for the 2018 ROADEF/EURO challenge glass cutting problem is actually also very competitive compared to other dedicated algorithms for guillotine Packing Problems from the literature, and is even able to return state-of-the-art solutions on several variants. Its performances seem to rely on two key components: a branching scheme which limits symmetry issues; and a tree search algorithm fully exploiting guide functions which make it possible to compare nodes at different levels of the search tree.

In addition to effectiveness, the choice of a tree search algorithm makes the algorithm attractive for problems with additional side constraints. Indeed, new constraints are likely to reduce the size of the search tree.

The algorithm is implemented in a new software package intended for researchers in Packing Problems to develop new branching schemes for other variants, for researchers in Artificial Intelligence to experiment new tree search algorithms, and for OR practitioners to quickly develop efficient algorithms implementing several business-specific constraints.

Future research will focus on developing algorithms for Cutting Stock and Variable-sized Bin Packing Problems, as well as branching schemes to generate another kind of patterns such as non-guillotine or non-staged guillotine ones.

\bibliographystyle{elsarticle-num-names}
\bibliography{main}

\begin{thebibliography}{56}
\providecommand{\natexlab}[1]{#1}
\providecommand{\url}[1]{\texttt{#1}}
\providecommand{\urlprefix}{URL }
\expandafter\ifx\csname urlstyle\endcsname\relax
  \providecommand{\doi}[1]{doi:\discretionary{}{}{}#1}\else
  \providecommand{\doi}[1]{doi:\discretionary{}{}{}\begingroup
  \urlstyle{rm}\url{#1}\endgroup}\fi
\providecommand{\bibinfo}[2]{#2}

\bibitem[{Libralesso and Fontan(2020)}]{libralesso_anytime_2020}
\bibinfo{author}{L.~Libralesso}, \bibinfo{author}{F.~Fontan},
  \bibinfo{title}{An anytime tree search algorithm for the 2018 {ROADEF}/{EURO}
  challenge glass cutting problem}, \bibinfo{journal}{arXiv:2004.00963 [cs]}
  \urlprefix\url{http://arxiv.org/abs/2004.00963}, \bibinfo{note}{arXiv:
  2004.00963}.

\bibitem[{Gilmore and Gomory(1965)}]{gilmore_multistage_1965}
\bibinfo{author}{P.~C. Gilmore}, \bibinfo{author}{R.~E. Gomory},
  \bibinfo{title}{Multistage {Cutting} {Stock} {Problems} of {Two} and {More}
  {Dimensions}}, \bibinfo{journal}{Operations Research}
  \bibinfo{volume}{13}~(\bibinfo{number}{1}) (\bibinfo{year}{1965})
  \bibinfo{pages}{94--120}, ISSN \bibinfo{issn}{0030-364X},
  \doi{\bibinfo{doi}{10.1287/opre.13.1.94}},
  \urlprefix\url{https://pubsonline.informs.org/doi/abs/10.1287/opre.13.1.94}.

\bibitem[{Velasco and Uchoa(2019)}]{velasco_improved_2019}
\bibinfo{author}{A.~S. Velasco}, \bibinfo{author}{E.~Uchoa},
  \bibinfo{title}{Improved state space relaxation for constrained
  two-dimensional guillotine cutting problems}, \bibinfo{journal}{European
  Journal of Operational Research} \bibinfo{volume}{272}~(\bibinfo{number}{1})
  (\bibinfo{year}{2019}) \bibinfo{pages}{106--120}, ISSN
  \bibinfo{issn}{0377-2217}, \doi{\bibinfo{doi}{10.1016/j.ejor.2018.06.016}},
  \urlprefix\url{http://www.sciencedirect.com/science/article/pii/S0377221718305393}.

\bibitem[{Wei et~al.(2014)Wei, Tian, Zhu, and Lim}]{wei_block-based_2014}
\bibinfo{author}{L.~Wei}, \bibinfo{author}{T.~Tian}, \bibinfo{author}{W.~Zhu},
  \bibinfo{author}{A.~Lim}, \bibinfo{title}{A block-based layer building
  approach for the {2D} guillotine strip packing problem},
  \bibinfo{journal}{European Journal of Operational Research}
  \bibinfo{volume}{239}~(\bibinfo{number}{1}) (\bibinfo{year}{2014})
  \bibinfo{pages}{58--69}, ISSN \bibinfo{issn}{0377-2217},
  \doi{\bibinfo{doi}{10.1016/j.ejor.2014.04.020}},
  \urlprefix\url{http://www.sciencedirect.com/science/article/pii/S0377221714003464}.

\bibitem[{Charalambous and Fleszar(2011)}]{charalambous_constructive_2011}
\bibinfo{author}{C.~Charalambous}, \bibinfo{author}{K.~Fleszar},
  \bibinfo{title}{A constructive bin-oriented heuristic for the two-dimensional
  bin packing problem with guillotine cuts}, \bibinfo{journal}{Computers \&
  Operations Research} \bibinfo{volume}{38}~(\bibinfo{number}{10})
  (\bibinfo{year}{2011}) \bibinfo{pages}{1443--1451}, ISSN
  \bibinfo{issn}{0305-0548}, \doi{\bibinfo{doi}{10.1016/j.cor.2010.12.013}},
  \urlprefix\url{http://www.sciencedirect.com/science/article/pii/S0305054810003096}.

\bibitem[{Fleszar(2013)}]{fleszar_three_2013}
\bibinfo{author}{K.~Fleszar}, \bibinfo{title}{Three insertion heuristics and a
  justification improvement heuristic for two-dimensional bin packing with
  guillotine cuts}, \bibinfo{journal}{Computers \& Operations Research}
  \bibinfo{volume}{40}~(\bibinfo{number}{1}) (\bibinfo{year}{2013})
  \bibinfo{pages}{463--474}, ISSN \bibinfo{issn}{0305-0548},
  \doi{\bibinfo{doi}{10.1016/j.cor.2012.07.016}},
  \urlprefix\url{http://www.sciencedirect.com/science/article/pii/S0305054812001621}.

\bibitem[{Cui et~al.(2018)Cui, Yao, and Zhang}]{cui_applying_2018}
\bibinfo{author}{Y.-P. Cui}, \bibinfo{author}{Y.~Yao},
  \bibinfo{author}{D.~Zhang}, \bibinfo{title}{Applying triple-block patterns in
  solving the two-dimensional bin packing problem}, \bibinfo{journal}{Journal
  of the Operational Research Society}
  \bibinfo{volume}{69}~(\bibinfo{number}{3}) (\bibinfo{year}{2018})
  \bibinfo{pages}{402--415}, ISSN \bibinfo{issn}{0160-5682},
  \doi{\bibinfo{doi}{10.1057/s41274-016-0148-5}},
  \urlprefix\url{https://orsociety.tandfonline.com/doi/abs/10.1057/s41274-016-0148-5}.

\bibitem[{Lodi and Monaci(2003)}]{lodi_integer_2003}
\bibinfo{author}{A.~Lodi}, \bibinfo{author}{M.~Monaci}, \bibinfo{title}{Integer
  linear programming models for 2-staged two-dimensional {Knapsack} problems},
  \bibinfo{journal}{Mathematical Programming}
  \bibinfo{volume}{94}~(\bibinfo{number}{2}) (\bibinfo{year}{2003})
  \bibinfo{pages}{257--278}, ISSN \bibinfo{issn}{1436-4646},
  \doi{\bibinfo{doi}{10.1007/s10107-002-0319-9}},
  \urlprefix\url{https://doi.org/10.1007/s10107-002-0319-9}.

\bibitem[{do~Nascimento et~al.(2019)do~Nascimento, de~Queiroz, and
  Junqueira}]{do_nascimento_mip-cp_2019}
\bibinfo{author}{O.~X. do~Nascimento}, \bibinfo{author}{T.~A. de~Queiroz},
  \bibinfo{author}{L.~Junqueira}, \bibinfo{title}{A {MIP}-{CP} based approach
  for two- and three-dimensional cutting problems with staged guillotine cuts},
  \bibinfo{journal}{Annals of Operations Research} ISSN
  \bibinfo{issn}{1572-9338}, \doi{\bibinfo{doi}{10.1007/s10479-019-03466-x}},
  \urlprefix\url{https://doi.org/10.1007/s10479-019-03466-x}.

\bibitem[{Bortfeldt and Winter(2009)}]{bortfeldt_genetic_2009}
\bibinfo{author}{A.~Bortfeldt}, \bibinfo{author}{T.~Winter}, \bibinfo{title}{A
  genetic algorithm for the two-dimensional knapsack problem with rectangular
  pieces}, \bibinfo{journal}{International Transactions in Operational
  Research} \bibinfo{volume}{16}~(\bibinfo{number}{6}) (\bibinfo{year}{2009})
  \bibinfo{pages}{685--713}, ISSN \bibinfo{issn}{1475-3995},
  \doi{\bibinfo{doi}{10.1111/j.1475-3995.2009.00701.x}},
  \urlprefix\url{https://onlinelibrary.wiley.com/doi/abs/10.1111/j.1475-3995.2009.00701.x}.

\bibitem[{Alvelos et~al.(2009)Alvelos, Chan, Vilaça, Gomes, Silva, and
  Carvalho}]{alvelos_sequence_2009}
\bibinfo{author}{F.~Alvelos}, \bibinfo{author}{T.~M. Chan},
  \bibinfo{author}{P.~Vilaça}, \bibinfo{author}{T.~Gomes},
  \bibinfo{author}{E.~Silva}, \bibinfo{author}{J.~M. V.~d. Carvalho},
  \bibinfo{title}{Sequence based heuristics for two-dimensional bin packing
  problems}, \bibinfo{journal}{Engineering Optimization}
  \bibinfo{volume}{41}~(\bibinfo{number}{8}) (\bibinfo{year}{2009})
  \bibinfo{pages}{773--791}, ISSN \bibinfo{issn}{0305-215X},
  \doi{\bibinfo{doi}{10.1080/03052150902835960}},
  \urlprefix\url{https://doi.org/10.1080/03052150902835960}.

\bibitem[{Silva et~al.(2010)Silva, Alvelos, and Valério~de
  Carvalho}]{silva_integer_2010}
\bibinfo{author}{E.~Silva}, \bibinfo{author}{F.~Alvelos},
  \bibinfo{author}{J.~M. Valério~de Carvalho}, \bibinfo{title}{An integer
  programming model for two- and three-stage two-dimensional cutting stock
  problems}, \bibinfo{journal}{European Journal of Operational Research}
  \bibinfo{volume}{205}~(\bibinfo{number}{3}) (\bibinfo{year}{2010})
  \bibinfo{pages}{699--708}, ISSN \bibinfo{issn}{0377-2217},
  \doi{\bibinfo{doi}{10.1016/j.ejor.2010.01.039}},
  \urlprefix\url{http://www.sciencedirect.com/science/article/pii/S0377221710000731}.

\bibitem[{Furini et~al.(2016)Furini, Malaguti, and
  Thomopulos}]{furini_modeling_2016}
\bibinfo{author}{F.~Furini}, \bibinfo{author}{E.~Malaguti},
  \bibinfo{author}{D.~Thomopulos}, \bibinfo{title}{Modeling {Two}-{Dimensional}
  {Guillotine} {Cutting} {Problems} via {Integer} {Programming}},
  \bibinfo{journal}{INFORMS Journal on Computing}
  \bibinfo{volume}{28}~(\bibinfo{number}{4}) (\bibinfo{year}{2016})
  \bibinfo{pages}{736--751}, ISSN \bibinfo{issn}{1091-9856},
  \doi{\bibinfo{doi}{10.1287/ijoc.2016.0710}},
  \urlprefix\url{https://pubsonline.informs.org/doi/10.1287/ijoc.2016.0710}.

\bibitem[{Lodi et~al.(2004)Lodi, Martello, and Vigo}]{lodi_models_2004}
\bibinfo{author}{A.~Lodi}, \bibinfo{author}{S.~Martello},
  \bibinfo{author}{D.~Vigo}, \bibinfo{title}{Models and {Bounds} for
  {Two}-{Dimensional} {Level} {Packing} {Problems}}, \bibinfo{journal}{Journal
  of Combinatorial Optimization} \bibinfo{volume}{8}~(\bibinfo{number}{3})
  (\bibinfo{year}{2004}) \bibinfo{pages}{363--379}, ISSN
  \bibinfo{issn}{1573-2886},
  \doi{\bibinfo{doi}{10.1023/B:JOCO.0000038915.62826.79}},
  \urlprefix\url{https://doi.org/10.1023/B:JOCO.0000038915.62826.79}.

\bibitem[{Nepomuceno et~al.(2008)Nepomuceno, Pinheiro, and
  Coelho}]{nepomuceno_hybrid_2008}
\bibinfo{author}{N.~Nepomuceno}, \bibinfo{author}{P.~Pinheiro},
  \bibinfo{author}{A.~L.~V. Coelho}, \bibinfo{title}{A {Hybrid} {Optimization}
  {Framework} for {Cutting} and {Packing} {Problems}}, in:
  \bibinfo{editor}{C.~Cotta}, \bibinfo{editor}{J.~van Hemert} (Eds.),
  \bibinfo{booktitle}{Recent {Advances} in {Evolutionary} {Computation} for
  {Combinatorial} {Optimization}}, Studies in {Computational} {Intelligence},
  \bibinfo{publisher}{Springer}, \bibinfo{address}{Berlin, Heidelberg}, ISBN
  \bibinfo{isbn}{978-3-540-70807-0}, \bibinfo{pages}{87--99},
  \doi{\bibinfo{doi}{10.1007/978-3-540-70807-0_6}},
  \urlprefix\url{https://doi.org/10.1007/978-3-540-70807-0_6},
  \bibinfo{year}{2008}.

\bibitem[{Akeb et~al.(2009)Akeb, Hifi, and M’Hallah}]{akeb_beam_2009}
\bibinfo{author}{H.~Akeb}, \bibinfo{author}{M.~Hifi},
  \bibinfo{author}{R.~M’Hallah}, \bibinfo{title}{A beam search algorithm for
  the circular packing problem}, \bibinfo{journal}{Computers \& Operations
  Research} \bibinfo{volume}{36}~(\bibinfo{number}{5}) (\bibinfo{year}{2009})
  \bibinfo{pages}{1513--1528}, ISSN \bibinfo{issn}{0305-0548},
  \doi{\bibinfo{doi}{10.1016/j.cor.2008.02.003}},
  \urlprefix\url{http://www.sciencedirect.com/science/article/pii/S0305054808000269}.

\bibitem[{Hifi and M'Hallah(2009)}]{hifi_beam_2009}
\bibinfo{author}{M.~Hifi}, \bibinfo{author}{R.~M'Hallah}, \bibinfo{title}{Beam
  search and non-linear programming tools for the circular packing problem},
  \bibinfo{journal}{International Journal of Mathematics in Operational
  Research} \bibinfo{volume}{1}~(\bibinfo{number}{4}) (\bibinfo{year}{2009})
  \bibinfo{pages}{476--503}, ISSN \bibinfo{issn}{1757-5850},
  \doi{\bibinfo{doi}{10.1504/IJMOR.2009.026278}},
  \urlprefix\url{https://www.inderscienceonline.com/doi/abs/10.1504/IJMOR.2009.026278}.

\bibitem[{Akeb et~al.(2010)Akeb, Hifi, and M'Hallah}]{akeb_adaptive_2010}
\bibinfo{author}{H.~Akeb}, \bibinfo{author}{M.~Hifi},
  \bibinfo{author}{R.~M'Hallah}, \bibinfo{title}{Adaptive beam search lookahead
  algorithms for the circular packing problem}, \bibinfo{journal}{International
  Transactions in Operational Research}
  \bibinfo{volume}{17}~(\bibinfo{number}{5}) (\bibinfo{year}{2010})
  \bibinfo{pages}{553--575}, ISSN \bibinfo{issn}{1475-3995},
  \doi{\bibinfo{doi}{10.1111/j.1475-3995.2009.00745.x}},
  \urlprefix\url{https://onlinelibrary.wiley.com/doi/abs/10.1111/j.1475-3995.2009.00745.x}.

\bibitem[{Akeb et~al.(2011)Akeb, Hifi, and Negre}]{akeb_augmented_2011}
\bibinfo{author}{H.~Akeb}, \bibinfo{author}{M.~Hifi},
  \bibinfo{author}{S.~Negre}, \bibinfo{title}{An augmented beam search-based
  algorithm for the circular open dimension problem},
  \bibinfo{journal}{Computers \& Industrial Engineering}
  \bibinfo{volume}{61}~(\bibinfo{number}{2}) (\bibinfo{year}{2011})
  \bibinfo{pages}{373--381}, ISSN \bibinfo{issn}{0360-8352},
  \doi{\bibinfo{doi}{10.1016/j.cie.2011.02.009}},
  \urlprefix\url{http://www.sciencedirect.com/science/article/pii/S0360835211000647}.

\bibitem[{Bennell and Song(2010)}]{bennell_beam_2010}
\bibinfo{author}{J.~A. Bennell}, \bibinfo{author}{X.~Song}, \bibinfo{title}{A
  beam search implementation for the irregular shape packing problem},
  \bibinfo{journal}{Journal of Heuristics}
  \bibinfo{volume}{16}~(\bibinfo{number}{2}) (\bibinfo{year}{2010})
  \bibinfo{pages}{167--188}, ISSN \bibinfo{issn}{1572-9397},
  \doi{\bibinfo{doi}{10.1007/s10732-008-9095-x}},
  \urlprefix\url{https://doi.org/10.1007/s10732-008-9095-x}.

\bibitem[{Bennell et~al.(2018)Bennell, Cabo, and
  Martínez-Sykora}]{bennell_beam_2018}
\bibinfo{author}{J.~A. Bennell}, \bibinfo{author}{M.~Cabo},
  \bibinfo{author}{A.~Martínez-Sykora}, \bibinfo{title}{A beam search approach
  to solve the convex irregular bin packing problem with guillotine cuts},
  \bibinfo{journal}{European Journal of Operational Research}
  \bibinfo{volume}{270}~(\bibinfo{number}{1}) (\bibinfo{year}{2018})
  \bibinfo{pages}{89--102}, ISSN \bibinfo{issn}{0377-2217},
  \doi{\bibinfo{doi}{10.1016/j.ejor.2018.03.029}},
  \urlprefix\url{http://www.sciencedirect.com/science/article/pii/S0377221718302571}.

\bibitem[{Wang et~al.(2013)Wang, Lim, and Zhu}]{wang_multi-round_2013}
\bibinfo{author}{N.~Wang}, \bibinfo{author}{A.~Lim}, \bibinfo{author}{W.~Zhu},
  \bibinfo{title}{A multi-round partial beam search approach for the single
  container loading problem with shipment priority},
  \bibinfo{journal}{International Journal of Production Economics}
  \bibinfo{volume}{145}~(\bibinfo{number}{2}) (\bibinfo{year}{2013})
  \bibinfo{pages}{531--540}, ISSN \bibinfo{issn}{0925-5273},
  \doi{\bibinfo{doi}{10.1016/j.ijpe.2013.04.028}},
  \urlprefix\url{http://www.sciencedirect.com/science/article/pii/S0925527313001928}.

\bibitem[{Araya and Riff(2014)}]{araya_beam_2014}
\bibinfo{author}{I.~Araya}, \bibinfo{author}{M.~C. Riff}, \bibinfo{title}{A
  beam search approach to the container loading problem},
  \bibinfo{journal}{Computers \& Operations Research} \bibinfo{volume}{43}
  (\bibinfo{year}{2014}) \bibinfo{pages}{100--107}, ISSN
  \bibinfo{issn}{0305-0548}, \doi{\bibinfo{doi}{10.1016/j.cor.2013.09.003}},
  \urlprefix\url{http://www.sciencedirect.com/science/article/pii/S0305054813002530}.

\bibitem[{Araya et~al.(2020)Araya, Moyano, and Sanchez}]{araya_beam_2020}
\bibinfo{author}{I.~Araya}, \bibinfo{author}{M.~Moyano},
  \bibinfo{author}{C.~Sanchez}, \bibinfo{title}{A beam search algorithm for the
  biobjective container loading problem}, \bibinfo{journal}{European Journal of
  Operational Research} ISSN \bibinfo{issn}{0377-2217},
  \doi{\bibinfo{doi}{10.1016/j.ejor.2020.03.040}},
  \urlprefix\url{http://www.sciencedirect.com/science/article/pii/S037722172030254X}.

\bibitem[{Hifi et~al.(2012)Hifi, Negre, Ouafi, and Saadi}]{hifi_parallel_2012}
\bibinfo{author}{M.~Hifi}, \bibinfo{author}{S.~Negre},
  \bibinfo{author}{R.~Ouafi}, \bibinfo{author}{T.~Saadi}, \bibinfo{title}{A
  parallel algorithm for constrained two-staged two-dimensional cutting
  problems}, \bibinfo{journal}{Computers \& Industrial Engineering}
  \bibinfo{volume}{62}~(\bibinfo{number}{1}) (\bibinfo{year}{2012})
  \bibinfo{pages}{177--189}, ISSN \bibinfo{issn}{0360-8352},
  \doi{\bibinfo{doi}{10.1016/j.cie.2011.09.005}},
  \urlprefix\url{http://www.sciencedirect.com/science/article/pii/S0360835211002737}.

\bibitem[{Christofides and Whitlock(1977)}]{christofides_algorithm_1977}
\bibinfo{author}{N.~Christofides}, \bibinfo{author}{C.~Whitlock},
  \bibinfo{title}{An {Algorithm} for {Two}-{Dimensional} {Cutting} {Problems}},
  \bibinfo{journal}{Operations Research}
  \bibinfo{volume}{25}~(\bibinfo{number}{1}) (\bibinfo{year}{1977})
  \bibinfo{pages}{30--44}, ISSN \bibinfo{issn}{0030-364X},
  \doi{\bibinfo{doi}{10.1287/opre.25.1.30}},
  \urlprefix\url{https://pubsonline.informs.org/doi/10.1287/opre.25.1.30}.

\bibitem[{Wang(1983)}]{wang_two_1983}
\bibinfo{author}{P.~Y. Wang}, \bibinfo{title}{Two {Algorithms} for
  {Constrained} {Two}-{Dimensional} {Cutting} {Stock} {Problems}},
  \bibinfo{journal}{Operations Research}
  \bibinfo{volume}{31}~(\bibinfo{number}{3}) (\bibinfo{year}{1983})
  \bibinfo{pages}{573--586}, ISSN \bibinfo{issn}{0030-364X},
  \doi{\bibinfo{doi}{10.1287/opre.31.3.573}},
  \urlprefix\url{https://pubsonline.informs.org/doi/abs/10.1287/opre.31.3.573}.

\bibitem[{Oliveira and Ferreira(1990)}]{oliveira_improved_1990}
\bibinfo{author}{J.~Oliveira}, \bibinfo{author}{J.~Ferreira},
  \bibinfo{title}{An improved version of {Wang}'s algorithm for two-dimensional
  cutting problems}, \bibinfo{journal}{European Journal of Operational
  Research} \bibinfo{volume}{44}~(\bibinfo{number}{2}) (\bibinfo{year}{1990})
  \bibinfo{pages}{256--266}, ISSN \bibinfo{issn}{0377-2217},
  \doi{\bibinfo{doi}{10.1016/0377-2217(90)90361-E}},
  \urlprefix\url{http://www.sciencedirect.com/science/article/pii/037722179090361E}.

\bibitem[{Tschöke and Holthöfer(1995)}]{tschoke_new_1995}
\bibinfo{author}{S.~Tschöke}, \bibinfo{author}{N.~Holthöfer},
  \bibinfo{title}{A new parallel approach to the constrained two-dimensional
  cutting stock problem}, in: \bibinfo{editor}{A.~Ferreira},
  \bibinfo{editor}{J.~Rolim} (Eds.), \bibinfo{booktitle}{Parallel {Algorithms}
  for {Irregularly} {Structured} {Problems}}, Lecture {Notes} in {Computer}
  {Science}, \bibinfo{publisher}{Springer}, \bibinfo{address}{Berlin,
  Heidelberg}, ISBN \bibinfo{isbn}{978-3-540-44915-7},
  \bibinfo{pages}{285--300}, \doi{\bibinfo{doi}{10.1007/3-540-60321-2_24}},
  \bibinfo{year}{1995}.

\bibitem[{Fekete and Schepers(1997)}]{fekete_new_1997}
\bibinfo{author}{S.~P. Fekete}, \bibinfo{author}{J.~Schepers},
  \bibinfo{title}{A new exact algorithm for general orthogonal d-dimensional
  knapsack problems}, in: \bibinfo{editor}{R.~Burkard},
  \bibinfo{editor}{G.~Woeginger} (Eds.), \bibinfo{booktitle}{Algorithms —
  {ESA} '97}, Lecture {Notes} in {Computer} {Science},
  \bibinfo{publisher}{Springer}, \bibinfo{address}{Berlin, Heidelberg}, ISBN
  \bibinfo{isbn}{978-3-540-69536-3}, \bibinfo{pages}{144--156},
  \doi{\bibinfo{doi}{10.1007/3-540-63397-9_12}}, \bibinfo{year}{1997}.

\bibitem[{Fayard et~al.(1998)Fayard, Hifi, and
  Zissimopoulos}]{fayard_efficient_1998}
\bibinfo{author}{D.~Fayard}, \bibinfo{author}{M.~Hifi},
  \bibinfo{author}{V.~Zissimopoulos}, \bibinfo{title}{An efficient approach for
  large-scale two-dimensional guillotine cutting stock problems},
  \bibinfo{journal}{Journal of the Operational Research Society}
  \bibinfo{volume}{49}~(\bibinfo{number}{12}) (\bibinfo{year}{1998})
  \bibinfo{pages}{1270--1277}, ISSN \bibinfo{issn}{0160-5682},
  \doi{\bibinfo{doi}{10.1057/palgrave.jors.2600638}},
  \urlprefix\url{https://doi.org/10.1057/palgrave.jors.2600638}.

\bibitem[{Hifi(1997)}]{hifi_improvement_1997}
\bibinfo{author}{M.~Hifi}, \bibinfo{title}{An improvement of viswanathan and
  bagchi's exact algorithm for constrained two-dimensional cutting stock},
  \bibinfo{journal}{Computers \& Operations Research}
  \bibinfo{volume}{24}~(\bibinfo{number}{8}) (\bibinfo{year}{1997})
  \bibinfo{pages}{727--736}, ISSN \bibinfo{issn}{0305-0548},
  \doi{\bibinfo{doi}{10.1016/S0305-0548(96)00095-0}},
  \urlprefix\url{http://www.sciencedirect.com/science/article/pii/S0305054896000950}.

\bibitem[{Cung et~al.(2000)Cung, Hifi, and Cun}]{cung_constrained_2000}
\bibinfo{author}{V.-D. Cung}, \bibinfo{author}{M.~Hifi}, \bibinfo{author}{B.~L.
  Cun}, \bibinfo{title}{Constrained two-dimensional cutting stock problems a
  best-first branch-and-bound algorithm}, \bibinfo{journal}{International
  Transactions in Operational Research}
  \bibinfo{volume}{7}~(\bibinfo{number}{3}) (\bibinfo{year}{2000})
  \bibinfo{pages}{185--210}, ISSN \bibinfo{issn}{1475-3995},
  \doi{\bibinfo{doi}{10.1111/j.1475-3995.2000.tb00194.x}},
  \urlprefix\url{https://onlinelibrary.wiley.com/doi/abs/10.1111/j.1475-3995.2000.tb00194.x}.

\bibitem[{Berkey and Wang(1987)}]{berkey_two-dimensional_1987}
\bibinfo{author}{J.~O. Berkey}, \bibinfo{author}{P.~Y. Wang},
  \bibinfo{title}{Two-{Dimensional} {Finite} {Bin}-{Packing} {Algorithms}},
  \bibinfo{journal}{Journal of the Operational Research Society}
  \bibinfo{volume}{38}~(\bibinfo{number}{5}) (\bibinfo{year}{1987})
  \bibinfo{pages}{423--429}, ISSN \bibinfo{issn}{1476-9360},
  \doi{\bibinfo{doi}{10.1057/jors.1987.70}},
  \urlprefix\url{https://doi.org/10.1057/jors.1987.70}.

\bibitem[{Martello and Vigo(1998)}]{martello_exact_1998}
\bibinfo{author}{S.~Martello}, \bibinfo{author}{D.~Vigo}, \bibinfo{title}{Exact
  {Solution} of the {Two}-{Dimensional} {Finite} {Bin} {Packing} {Problem}},
  \bibinfo{journal}{Management Science}
  \bibinfo{volume}{44}~(\bibinfo{number}{3}) (\bibinfo{year}{1998})
  \bibinfo{pages}{388--399}, ISSN \bibinfo{issn}{0025-1909},
  \doi{\bibinfo{doi}{10.1287/mnsc.44.3.388}},
  \urlprefix\url{https://pubsonline.informs.org/doi/abs/10.1287/mnsc.44.3.388}.

\bibitem[{Beasley(1985)}]{beasley_algorithms_1985}
\bibinfo{author}{J.~E. Beasley}, \bibinfo{title}{Algorithms for {Unconstrained}
  {Two}-{Dimensional} {Guillotine} {Cutting}}, \bibinfo{journal}{Journal of the
  Operational Research Society} \bibinfo{volume}{36}~(\bibinfo{number}{4})
  (\bibinfo{year}{1985}) \bibinfo{pages}{297--306}, ISSN
  \bibinfo{issn}{1476-9360}, \doi{\bibinfo{doi}{10.1057/jors.1985.51}},
  \urlprefix\url{https://doi.org/10.1057/jors.1985.51}.

\bibitem[{Kröger(1995)}]{kroger_guillotineable_1995}
\bibinfo{author}{B.~Kröger}, \bibinfo{title}{Guillotineable bin packing: {A}
  genetic approach}, \bibinfo{journal}{European Journal of Operational
  Research} \bibinfo{volume}{84}~(\bibinfo{number}{3}) (\bibinfo{year}{1995})
  \bibinfo{pages}{645--661}, ISSN \bibinfo{issn}{0377-2217},
  \doi{\bibinfo{doi}{10.1016/0377-2217(95)00029-P}},
  \urlprefix\url{http://www.sciencedirect.com/science/article/pii/037722179500029P}.

\bibitem[{Hopper(2000)}]{hopper_two-dimensional_2000}
\bibinfo{author}{E.~Hopper}, \bibinfo{title}{Two-dimensional packing utilising
  evolutionary algorithms and other meta-heuristic methods},
  \bibinfo{type}{{PhD} {Thesis}}, \bibinfo{school}{University of Wales.
  Cardiff}, \bibinfo{year}{2000}.

\bibitem[{Hopper and Turton(2001)}]{hopper_empirical_2001}
\bibinfo{author}{E.~Hopper}, \bibinfo{author}{B.~C.~H. Turton},
  \bibinfo{title}{An empirical investigation of meta-heuristic and heuristic
  algorithms for a {2D} packing problem}, \bibinfo{journal}{European Journal of
  Operational Research} \bibinfo{volume}{128}~(\bibinfo{number}{1})
  (\bibinfo{year}{2001}) \bibinfo{pages}{34--57}, ISSN
  \bibinfo{issn}{0377-2217},
  \doi{\bibinfo{doi}{10.1016/S0377-2217(99)00357-4}},
  \urlprefix\url{http://www.sciencedirect.com/science/article/pii/S0377221799003574}.

\bibitem[{Alvarez-Valdés et~al.(2002)Alvarez-Valdés, Parajón, and
  Tamarit}]{alvarez-valdes_tabu_2002}
\bibinfo{author}{R.~Alvarez-Valdés}, \bibinfo{author}{A.~Parajón},
  \bibinfo{author}{J.~M. Tamarit}, \bibinfo{title}{A tabu search algorithm for
  large-scale guillotine (un)constrained two-dimensional cutting problems},
  \bibinfo{journal}{Computers \& Operations Research}
  \bibinfo{volume}{29}~(\bibinfo{number}{7}) (\bibinfo{year}{2002})
  \bibinfo{pages}{925--947}, ISSN \bibinfo{issn}{0305-0548},
  \doi{\bibinfo{doi}{10.1016/S0305-0548(00)00095-2}},
  \urlprefix\url{http://www.sciencedirect.com/science/article/pii/S0305054800000952}.

\bibitem[{Morabito and Pureza(2010)}]{morabito_heuristic_2010}
\bibinfo{author}{R.~Morabito}, \bibinfo{author}{V.~Pureza}, \bibinfo{title}{A
  heuristic approach based on dynamic programming and and/or-graph search for
  the constrained two-dimensional guillotine cutting problem},
  \bibinfo{journal}{Annals of Operations Research}
  \bibinfo{volume}{179}~(\bibinfo{number}{1}) (\bibinfo{year}{2010})
  \bibinfo{pages}{297--315}, ISSN \bibinfo{issn}{1572-9338},
  \doi{\bibinfo{doi}{10.1007/s10479-008-0457-4}},
  \urlprefix\url{https://doi.org/10.1007/s10479-008-0457-4}.

\bibitem[{Cui and Zhao(2013)}]{cui_heuristic_2013-1}
\bibinfo{author}{Y.~Cui}, \bibinfo{author}{Z.~Zhao}, \bibinfo{title}{Heuristic
  for the rectangular two-dimensional single stock size cutting stock problem
  with two-staged patterns}, \bibinfo{journal}{European Journal of Operational
  Research} \bibinfo{volume}{231}~(\bibinfo{number}{2}) (\bibinfo{year}{2013})
  \bibinfo{pages}{288--298}, ISSN \bibinfo{issn}{0377-2217},
  \doi{\bibinfo{doi}{10.1016/j.ejor.2013.05.042}},
  \urlprefix\url{http://www.sciencedirect.com/science/article/pii/S0377221713004591}.

\bibitem[{Cui et~al.(2016)Cui, Yao, and Cui}]{cui_hybrid_2016}
\bibinfo{author}{Y.~Cui}, \bibinfo{author}{Y.~Yao}, \bibinfo{author}{Y.-P.
  Cui}, \bibinfo{title}{Hybrid approach for the two-dimensional bin packing
  problem with two-staged patterns}, \bibinfo{journal}{International
  Transactions in Operational Research}
  \bibinfo{volume}{23}~(\bibinfo{number}{3}) (\bibinfo{year}{2016})
  \bibinfo{pages}{539--549}, ISSN \bibinfo{issn}{1475-3995},
  \doi{\bibinfo{doi}{10.1111/itor.12188}},
  \urlprefix\url{https://onlinelibrary.wiley.com/doi/abs/10.1111/itor.12188}.

\bibitem[{Puchinger and Raidl(2007)}]{puchinger_models_2007}
\bibinfo{author}{J.~Puchinger}, \bibinfo{author}{G.~R. Raidl},
  \bibinfo{title}{Models and algorithms for three-stage two-dimensional bin
  packing}, \bibinfo{journal}{European Journal of Operational Research}
  \bibinfo{volume}{183}~(\bibinfo{number}{3}) (\bibinfo{year}{2007})
  \bibinfo{pages}{1304--1327}, ISSN \bibinfo{issn}{0377-2217},
  \doi{\bibinfo{doi}{10.1016/j.ejor.2005.11.064}},
  \urlprefix\url{http://www.sciencedirect.com/science/article/pii/S0377221706003067}.

\bibitem[{Alvelos et~al.(2014)Alvelos, Silva, and
  de~Carvalho}]{alvelos_hybrid_2014}
\bibinfo{author}{F.~Alvelos}, \bibinfo{author}{E.~Silva},
  \bibinfo{author}{J.~M.~V. de~Carvalho}, \bibinfo{title}{A {Hybrid}
  {Heuristic} {Based} on {Column} {Generation} for {Two}- and {Three}- {Stage}
  {Bin} {Packing} {Problems}}, in: \bibinfo{editor}{B.~Murgante},
  \bibinfo{editor}{S.~Misra}, \bibinfo{editor}{A.~M. A.~C. Rocha},
  \bibinfo{editor}{C.~Torre}, \bibinfo{editor}{J.~G. Rocha},
  \bibinfo{editor}{M.~I. Falcão}, \bibinfo{editor}{D.~Taniar},
  \bibinfo{editor}{B.~O. Apduhan}, \bibinfo{editor}{O.~Gervasi} (Eds.),
  \bibinfo{booktitle}{Computational {Science} and {Its} {Applications} –
  {ICCSA} 2014}, Lecture {Notes} in {Computer} {Science},
  \bibinfo{publisher}{Springer International Publishing},
  \bibinfo{address}{Cham}, ISBN \bibinfo{isbn}{978-3-319-09129-7},
  \bibinfo{pages}{211--226}, \doi{\bibinfo{doi}{10.1007/978-3-319-09129-7_16}},
  \bibinfo{year}{2014}.

\bibitem[{Cui et~al.(2015)Cui, Cui, Tang, and Hu}]{cui_heuristic_2015}
\bibinfo{author}{Y.-P. Cui}, \bibinfo{author}{Y.~Cui},
  \bibinfo{author}{T.~Tang}, \bibinfo{author}{W.~Hu}, \bibinfo{title}{Heuristic
  for constrained two-dimensional three-staged patterns},
  \bibinfo{journal}{Journal of the Operational Research Society}
  \bibinfo{volume}{66}~(\bibinfo{number}{4}) (\bibinfo{year}{2015})
  \bibinfo{pages}{647--656}, ISSN \bibinfo{issn}{0160-5682},
  \doi{\bibinfo{doi}{10.1057/jors.2014.33}},
  \urlprefix\url{https://orsociety.tandfonline.com/doi/abs/10.1057/jors.2014.33}.

\bibitem[{Alvarez-Valdes et~al.(2007)Alvarez-Valdes, Martí, Tamarit, and
  Parajón}]{alvarez-valdes_grasp_2007}
\bibinfo{author}{R.~Alvarez-Valdes}, \bibinfo{author}{R.~Martí},
  \bibinfo{author}{J.~M. Tamarit}, \bibinfo{author}{A.~Parajón},
  \bibinfo{title}{{GRASP} and {Path} {Relinking} for the {Two}-{Dimensional}
  {Two}-{Stage} {Cutting}-{Stock} {Problem}}, \bibinfo{journal}{INFORMS Journal
  on Computing} \bibinfo{volume}{19}~(\bibinfo{number}{2})
  (\bibinfo{year}{2007}) \bibinfo{pages}{261--272}, ISSN
  \bibinfo{issn}{1091-9856}, \doi{\bibinfo{doi}{10.1287/ijoc.1050.0169}},
  \urlprefix\url{https://pubsonline.informs.org/doi/abs/10.1287/ijoc.1050.0169}.

\bibitem[{Hifi et~al.(2008)Hifi, M'Hallah, and Saadi}]{hifi_algorithms_2008}
\bibinfo{author}{M.~Hifi}, \bibinfo{author}{R.~M'Hallah},
  \bibinfo{author}{T.~Saadi}, \bibinfo{title}{Algorithms for the {Constrained}
  {Two}-{Staged} {Two}-{Dimensional} {Cutting} {Problem}},
  \bibinfo{journal}{INFORMS Journal on Computing}
  \bibinfo{volume}{20}~(\bibinfo{number}{2}) (\bibinfo{year}{2008})
  \bibinfo{pages}{212--221}, ISSN \bibinfo{issn}{1091-9856},
  \doi{\bibinfo{doi}{10.1287/ijoc.1070.0233}},
  \urlprefix\url{https://pubsonline.informs.org/doi/10.1287/ijoc.1070.0233}.

\bibitem[{Hifi and Roucairol(2001)}]{hifi_approximate_2001}
\bibinfo{author}{M.~Hifi}, \bibinfo{author}{C.~Roucairol},
  \bibinfo{title}{Approximate and {Exact} {Algorithms} for {Constrained} ({Un})
  {Weighted} {Two}-dimensional {Two}-staged {Cutting} {Stock} {Problems}},
  \bibinfo{journal}{Journal of Combinatorial Optimization}
  \bibinfo{volume}{5}~(\bibinfo{number}{4}) (\bibinfo{year}{2001})
  \bibinfo{pages}{465--494}, ISSN \bibinfo{issn}{1573-2886},
  \doi{\bibinfo{doi}{10.1023/A:1011628809603}},
  \urlprefix\url{https://doi.org/10.1023/A:1011628809603}.

\bibitem[{Wei and Lim(2015)}]{wei_bidirectional_2015}
\bibinfo{author}{L.~Wei}, \bibinfo{author}{A.~Lim}, \bibinfo{title}{A
  bidirectional building approach for the {2D} constrained guillotine knapsack
  packing problem}, \bibinfo{journal}{European Journal of Operational Research}
  \bibinfo{volume}{242}~(\bibinfo{number}{1}) (\bibinfo{year}{2015})
  \bibinfo{pages}{63--71}, ISSN \bibinfo{issn}{0377-2217},
  \doi{\bibinfo{doi}{10.1016/j.ejor.2014.10.004}},
  \urlprefix\url{http://www.sciencedirect.com/science/article/pii/S037722171400808X}.

\bibitem[{Dolatabadi et~al.(2012)Dolatabadi, Lodi, and
  Monaci}]{dolatabadi_exact_2012}
\bibinfo{author}{M.~Dolatabadi}, \bibinfo{author}{A.~Lodi},
  \bibinfo{author}{M.~Monaci}, \bibinfo{title}{Exact algorithms for the
  two-dimensional guillotine knapsack}, \bibinfo{journal}{Computers \&
  Operations Research} \bibinfo{volume}{39}~(\bibinfo{number}{1})
  (\bibinfo{year}{2012}) \bibinfo{pages}{48--53}, ISSN
  \bibinfo{issn}{0305-0548}, \doi{\bibinfo{doi}{10.1016/j.cor.2010.12.018}},
  \urlprefix\url{http://www.sciencedirect.com/science/article/pii/S0305054811000190}.

\bibitem[{Cui et~al.(2008)Cui, Gu, and Hu}]{cui_algorithm_2008}
\bibinfo{author}{Y.~Cui}, \bibinfo{author}{T.~Gu}, \bibinfo{author}{W.~Hu},
  \bibinfo{title}{An algorithm for the constrained two-dimensional rectangular
  multiple identical large object placement problem},
  \bibinfo{journal}{Optimization Methods and Software}
  \bibinfo{volume}{23}~(\bibinfo{number}{3}) (\bibinfo{year}{2008})
  \bibinfo{pages}{375--393}, ISSN \bibinfo{issn}{1055-6788},
  \doi{\bibinfo{doi}{10.1080/10556780701617163}},
  \urlprefix\url{https://doi.org/10.1080/10556780701617163}.

\bibitem[{Cui et~al.(2017)Cui, Zhou, and Cui}]{cui_triple-solution_2017}
\bibinfo{author}{Y.-P. Cui}, \bibinfo{author}{Y.~Zhou},
  \bibinfo{author}{Y.~Cui}, \bibinfo{title}{Triple-solution approach for the
  strip packing problem with two-staged patterns}, \bibinfo{journal}{Journal of
  Combinatorial Optimization} \bibinfo{volume}{34}~(\bibinfo{number}{2})
  (\bibinfo{year}{2017}) \bibinfo{pages}{588--604}, ISSN
  \bibinfo{issn}{1573-2886}, \doi{\bibinfo{doi}{10.1007/s10878-016-0088-7}},
  \urlprefix\url{https://doi.org/10.1007/s10878-016-0088-7}.

\bibitem[{Cui et~al.(2013)Cui, Yang, and Chen}]{cui_heuristic_2013}
\bibinfo{author}{Y.~Cui}, \bibinfo{author}{L.~Yang}, \bibinfo{author}{Q.~Chen},
  \bibinfo{title}{Heuristic for the rectangular strip packing problem with
  rotation of items}, \bibinfo{journal}{Computers \& Operations Research}
  \bibinfo{volume}{40}~(\bibinfo{number}{4}) (\bibinfo{year}{2013})
  \bibinfo{pages}{1094--1099}, ISSN \bibinfo{issn}{0305-0548},
  \doi{\bibinfo{doi}{10.1016/j.cor.2012.11.020}},
  \urlprefix\url{http://www.sciencedirect.com/science/article/pii/S030505481200264X}.

\bibitem[{Bortfeldt and Jungmann(2012)}]{bortfeldt_tree_2012}
\bibinfo{author}{A.~Bortfeldt}, \bibinfo{author}{S.~Jungmann},
  \bibinfo{title}{A tree search algorithm for solving the multi-dimensional
  strip packing problem with guillotine cutting constraint},
  \bibinfo{journal}{Annals of Operations Research}
  \bibinfo{volume}{196}~(\bibinfo{number}{1}) (\bibinfo{year}{2012})
  \bibinfo{pages}{53--71}, ISSN \bibinfo{issn}{1572-9338},
  \doi{\bibinfo{doi}{10.1007/s10479-012-1084-7}},
  \urlprefix\url{https://doi.org/10.1007/s10479-012-1084-7}.

\bibitem[{Lodi et~al.(2017)Lodi, Monaci, and Pietrobuoni}]{lodi_partial_2017}
\bibinfo{author}{A.~Lodi}, \bibinfo{author}{M.~Monaci},
  \bibinfo{author}{E.~Pietrobuoni}, \bibinfo{title}{Partial enumeration
  algorithms for {Two}-{Dimensional} {Bin} {Packing} {Problem} with guillotine
  constraints}, \bibinfo{journal}{Discrete Applied Mathematics}
  \bibinfo{volume}{217} (\bibinfo{year}{2017}) \bibinfo{pages}{40--47}, ISSN
  \bibinfo{issn}{0166-218X}, \doi{\bibinfo{doi}{10.1016/j.dam.2015.09.012}},
  \urlprefix\url{http://www.sciencedirect.com/science/article/pii/S0166218X15004734}.

\end{thebibliography}

\end{document}